\documentclass{article}

\usepackage{PRIMEarxiv}

\usepackage{xcolor}
\usepackage{comment}
\usepackage[utf8]{inputenc} 
\usepackage[T1]{fontenc}    
\usepackage{hyperref}       
\usepackage{url}            
\usepackage{booktabs}       
\usepackage{amsfonts}       
\usepackage{nicefrac}       
\usepackage{microtype}      
\usepackage{lipsum}
\usepackage{multirow}
\usepackage{fancyhdr}       
\usepackage{graphicx}       
\graphicspath{{media/}}     

\title{Large Language Models in Healthcare and Medical Domain: A Review
}

\author{
  Zabir Al Nazi \\
  University of California, Riverside \\
  Riverside, CA \\
  \texttt{znazi002@ucr.edu} \\
   \And
  Wei Peng \\
  Stanford University \\
  Palo Alto, CA\\
  \texttt{wepeng@stanford.edu} \\
}

\begin{document}
\maketitle

\begin{abstract}
The deployment of large language models (LLMs) within the healthcare sector has sparked both enthusiasm and apprehension. These
models exhibit the remarkable capability to provide proficient responses to free-text queries, demonstrating a nuanced understanding
of professional medical knowledge. This comprehensive survey delves into the functionalities of existing LLMs designed for
healthcare applications, elucidating the trajectory of their development, starting from traditional Pretrained Language Models (PLMs)
to the present state of LLMs in healthcare sector. First, we explore the potential of LLMs to amplify the efficiency and effectiveness
of diverse healthcare applications, particularly focusing on clinical language understanding tasks. These tasks encompass a wide
spectrum, ranging from named entity recognition and relation extraction to natural language inference, multi-modal medical
applications, document classification, and question-answering. Additionally, we conduct an extensive comparison of the most recent
state-of-the-art LLMs in the healthcare domain, while also assessing the utilization of various open-source LLMs and highlighting
their significance in healthcare applications. Furthermore, we present the essential performance metrics employed to evaluate LLMs
in the biomedical domain, shedding light on their effectiveness and limitations. Finally, we summarize the prominent challenges and
constraints faced by large language models in the healthcare sector, offering a holistic perspective on their potential benefits and
shortcomings. This review provides a comprehensive exploration of the current landscape of LLMs in healthcare, addressing their
role in transforming medical applications and the areas that warrant further research and development.
\end{abstract}


\section{Introduction}
Deep Learning provides an intelligent way to understand human behaviors, emotions and human healthcare~\cite{shi2022multiscale,yu2023modality,li2021micro,hong2019characterizing}.
Recent developments in clinical language understanding have ushered in the potential for a paradigm shift in the healthcare sector. These advancements hold the promise of ushering in a new era characterized by the deployment of intelligent systems designed to bolster decision-making, expedite diagnostic processes, and elevate the quality of patient care. In essence, these systems have the capacity to serve as indispensable aids to healthcare professionals as they grapple with the ever-expanding body of medical knowledge, decipher intricate patient records, and formulate highly tailored treatment plans. This transformative potential has ignited considerable enthusiasm within the healthcare community \cite{he2023survey, wang2023large, yu2023leveraging}.

The immense value of lage language models (LLMs) lies in their ability to process and synthesize colossal volumes of medical literature, patient records, and the ever-expanding body of clinical research. Healthcare data \cite{peng2022learning,peng2023generating} is inherently complex, heterogeneous, and often overwhelming in scale. LLMs act as a powerful force multiplier,  aiding healthcare professionals struggling with information overload. By automating the analysis of medical texts, extracting crucial insights, and applying that knowledge, LLMs are poised to drive groundbreaking research and enhance patient care, significantly improving and contributing to the progression of the healthcare and medical domain.

Notably, this surge of enthusiasm is attributable, in part, to the exceptional performance of state-of-the-art large language models (LLMs) such as OpenAI's GPT-3.5, GPT-4 \cite{brown2020language, openai2023gpt4}, and Google's Bard. These models have exhibited remarkable proficiency in a wide spectrum of natural language understanding tasks, highlighting their pivotal role in healthcare. Their ability to comprehend and generate human-like text is poised to play a transformative role in healthcare practices, where effective communication and information processing are of paramount importance \cite{zhang2023one}.

The trajectory of natural language processing (NLP) has been characterized by a series of noteworthy milestones, with each development building upon the strengths and limitations of its predecessors. In its nascent stages, recurrent neural networks (RNNs) laid the foundation for contextual information retention in NLP tasks. However, their inherent limitations in capturing long-range dependencies became evident, thus necessitating a shift in the NLP paradigm.

The pivotal moment in NLP's evolution came with the introduction of Transformers, a groundbreaking architecture that addressed the challenge of capturing distant word relationships effectively. This innovation was a turning point, enabling more advanced NLP models. These advancements provided the impetus for the emergence of sophisticated language models like Llama 2 \cite{touvron2023llama} and GPT-4, which, underpinned by extensive training data, have elevated NLP to a level of understanding and text generation that closely approximates human-like language.

Within the healthcare domain, tailored adaptations of models like BERT, including BioBERT and ClinicalBERT \cite{lee2020biobert, huang2019clinicalbert}, were introduced to tackle the intricacies of clinical language. The introduction of these models addressed the unique challenges posed by medical text, which frequently features complex medical terminology, lexical ambiguity, and variable usage. However, introducing LLMs into the highly sensitive and regulated domain of healthcare demands careful consideration of ethics, privacy, and security. Patient data must be rigorously protected, while ensuring that LLMs don't perpetuate existing biases or lead to unintended harm. Nevertheless, the potential for LLMs to enhance healthcare practices, better patient outcomes, and spearhead innovative research avenues continues to stimulate ongoing investigation and growth in this rapidly evolving field.

As we navigate this dynamic field, our review aims to function as a comprehensive guide, offering insights to medical researchers and healthcare professionals seeking to optimize their research endeavors and clinical practices. We seek to provide a valuable resource for the judicious selection of LLMs tailored to specific clinical requirements. Our examination encompasses a detailed exploration of LLMs within the healthcare domain, elucidating their underlying technology, diverse healthcare applications, and facilitating discussions on critical topics such as fairness, bias mitigation, privacy, transparency, and ethical considerations. By highlighting these critical aspects, this review aims to illustrate the importance of integrating LLMs into healthcare in a manner that is not only effective but also ethical, fair, and equitable, ultimately fostering benefits for both patients and healthcare providers.

This review paper is organized into distinct sections that systematically address the integration, impact, and limitations of large language models (LLMs) in healthcare:
\begin{itemize}

\item Section \textbf{2} provides a foundational understanding of LLMs, covering their key architectures such as Transformers, foundational models, and multi-modal capabilities.

\item In section \textbf{3}, the focus shifts to the application of LLMs in healthcare, discussing their use cases and the metrics for assessing their performance within clinical settings.

\item Section \textbf{4} critically examines the challenges associated with LLMs in healthcare, including issues related to explainability, security, bias, and ethical considerations.

\item The paper concludes by summarizing the findings, highlighting the transformative potential of LLMs while acknowledging the need for careful implementation to navigate their limitations and ethical implications.
\end{itemize}

\section{ Review of Large Language Models }
\label{rev}

Large language models have emerged as a notable advancement in the field of natural language processing (NLP) and have attracted considerable interest in recent times \cite{petroni2019language, brown2020language}. These models exhibit notable attributes such as their considerable number of parameters, pre-training on vast collections of textual data, and fine-tuning for specific downstream objectives \cite{radford2018improving, chowdhery2022palm, touvron2023llama}. By leveraging these key characteristics, large language models demonstrate exceptional performance across a wide range of NLP tasks. This section presents a comprehensive discussion of the concept, architecture, and pioneering examples of large language models. Furthermore, we explore the pre-training methodology and the significance of transfer learning in facilitating these models to achieve exceptional performance across diverse tasks \cite{radford2019language}.

Large Language models, built upon the Transformer architecture, have been specifically engineered to enhance the efficiency of natural language data processing in comparison to earlier iterations. The Transformer architecture, as proposed by \cite{vaswani2017attention}, utilizes a self-attention mechanism to capture the contextual relationships between words in a sentence. This mechanism facilitates the model's ability to assign varying degrees of significance to distinct words during the prediction process, rendering it especially suitable for handling long-range dependencies in language.

The key aspects of large language models encompass their substantial magnitude \cite{fedus2022switch, du2022glam}, pre-training on vast text corpora \cite{wang2022pre, touvron2023llama}, and subsequent fine-tuning tailored towards specific tasks \cite{wei2021finetuned}. These models possess a substantial number of parameters, ranging from hundreds of millions to billions, which allows them to effectively capture intricate patterns and nuances within language. Pre-training is commonly conducted on diverse datasets devoid of task-specific annotations, enabling the model to acquire knowledge from a broad spectrum of linguistic instances and develop a comprehensive grasp of language. Following pre-training, the model undergoes a further fine-tuning process using smaller datasets that are appropriate to the task at hand. This allows the model to successfully adapt to and perform well on specific natural language processing (NLP) tasks.

The progression of natural language processing (NLP) has been characterized by a series of significant advancements. At the outset, recurrent neural networks (RNNs) facilitated the retention of context in natural language processing (NLP) tasks. Nevertheless, recurrent neural networks (RNNs) were found to have several shortcomings when it comes to effectively capturing long-range dependencies. The advent of Transformers has had a transformative impact by effectively addressing the challenge of capturing distant word relationships. Subsequently, large language models like Llama 2 \cite{touvron2023llama}, GPT-4 \cite{openai2023gpt4} emerged, powered by extensive training data, significantly advancing NLP capabilities in understanding and generating human-like text. This progression signifies a continuous cycle of innovation, with each stage building upon the strengths and limitations of its predecessor. In the subsequent section, we delineate significant phases of development within the continuum of progress in the landscape of natural language processing (NLP).

In the domain of healthcare, specialized adaptations of BERT, namely BioBERT \cite{lee2020biobert} and ClinicalBERT \cite{huang2019clinicalbert}, were introduced to address a variety of challenges in comprehending clinical language. GPT-3 (Generative Pre-trained Transformer 3), developed by OpenAI, is one of the largest language models to date, with 175 billion parameters \cite{brown2020language}. Recently, OpenAI introduced the GPT-3.5 and its successor, GPT-4 (OpenAI, 2023) \cite{openai2023gpt4}, alongside Google AI's Bard, both of which have emerged as cutting-edge Large Language Models (LLMs), displaying remarkable capabilities across diverse applications, including healthcare and medicine \cite{wang2023large}.

\subsection{Transformers}

The Transformers architecture, introduced in "Attention is All You Need," \cite{vaswani2017attention} has revolutionized natural language processing. The primary novelty of this model is its utilization of the self-attention mechanism, which allows for the assessment of the importance of input tokens by considering their relevance to the given task. In this setup, multiple attention heads work in parallel, allowing the model to focus on various aspects of the input whereas positional encoding conveys relative token positions. 
Given an input sequence $X$ of length $N$, the self-attention mechanism \cite{zhou2021refiner, al2021fibro, hao2021self} computes attention scores $A(i, j)$ between all token pairs $(i, j)$. Three learned matrices, Query $(Q)$, Key $(K)$, and Value $(V)$, are obtained by linear projections of $X$. 

\[ Attention(Q, K, V) = softmax(\frac{QK^T}{\sqrt{d_k}})V \]

Here, $d_k$ represents the dimension of key vectors. The softmax function normalizes scores. The output for each token is then computed as a weighted sum of value vectors for all tokens j. Multi-Head Attention extends this mechanism by computing multiple attention sets in parallel, concatenated and linearly transformed to form the final output.

Transformers consist of stacked encoder-decoder blocks, adapting to diverse tasks. The training occurs via unsupervised or semi-supervised learning on vast text corpora, using gradient-based optimization. Transformers have become foundational in natural language processing due to their capacity to handle sequential data, capture long-range dependencies, and adapt to various tasks with minimal fine-tuning. They extend beyond text, finding applications in healthcare, recommendation systems, image generation, and other domains.

\begin{figure*}

 \center

  \includegraphics[width=0.9\textwidth]{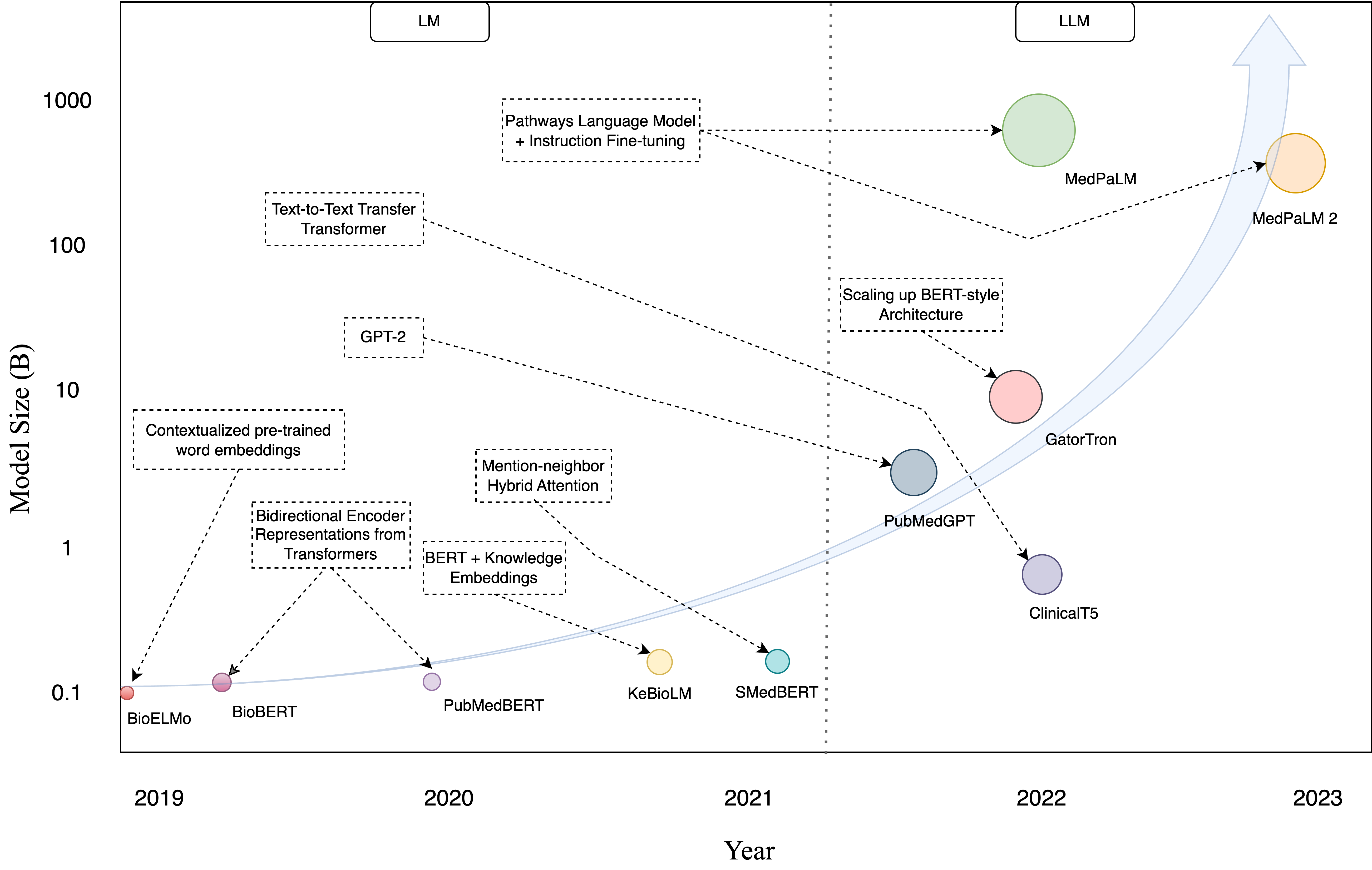}

  \caption{Scale of Medical Language Models: A Size Comparison} 

  \label{model_size}
\label{fig:scale_llm}
\end{figure*}

\subsection{Large Foundational Models}

The advent of Large Foundational Models, exemplified by GPT-3 (Brown et al., 2020) \cite{brown2020language} and Stable Diffusion (Rombach et al., 2022) \cite{rombach2022high}, ushers in a transformative era in the field of machine learning and generative artificial intelligence. Researchers have introduced the term "foundation model" to delineate machine learning models that undergo training on extensive, diverse, and unlabeled datasets, endowing them with the ability to adeptly tackle a broad spectrum of general tasks. These encompass tasks related to language comprehension, text and image generation, and natural language dialogue.

Large foundational models are massive AI architectures trained on extensive quantities of unlabeled data, predominantly employing self-supervised learning methods. This approach to training yields models of exceptional versatility, enabling them to excel across a wide spectrum of tasks, ranging from image classification and natural language processing to question-answering, consistently delivering outstanding levels of accuracy.

These models particularly shine in tasks demanding generative capabilities and human interaction, including the creation of marketing content or intricate artwork based on minimal prompts. Nevertheless, adapting and integrating these models into enterprise applications may present specific challenges \cite{rawte2023survey}.

\subsection{Multi-modal Language Models}

A Multi-Modal Large Language Model (MLLM) represents a groundbreaking advancement in the fields of artificial intelligence (AI) and natural language processing (NLP). In contrast to conventional language models focused solely on textual data, MLLMs possess the unique ability to process and generate content across multiple modalities, including text, images, audio, and video. This novel approach significantly expands the capabilities of AI applications, allowing machines not only to comprehend and generate text but also to interpret and integrate information from various sensory inputs. The integration of multiple modalities enables MLLMs to bridge the gap between human communication and machine understanding, making them versatile tools with the potential to transform diverse fields. This theoretical introduction highlights the transformative potential of MLLMs and their central role in pushing the boundaries of artificial intelligence, affecting areas such as image and speech recognition, content generation, and interactive AI applications \cite{yin2023survey}.

\begin{figure*}

 \center

  \includegraphics[width=0.80\textwidth]{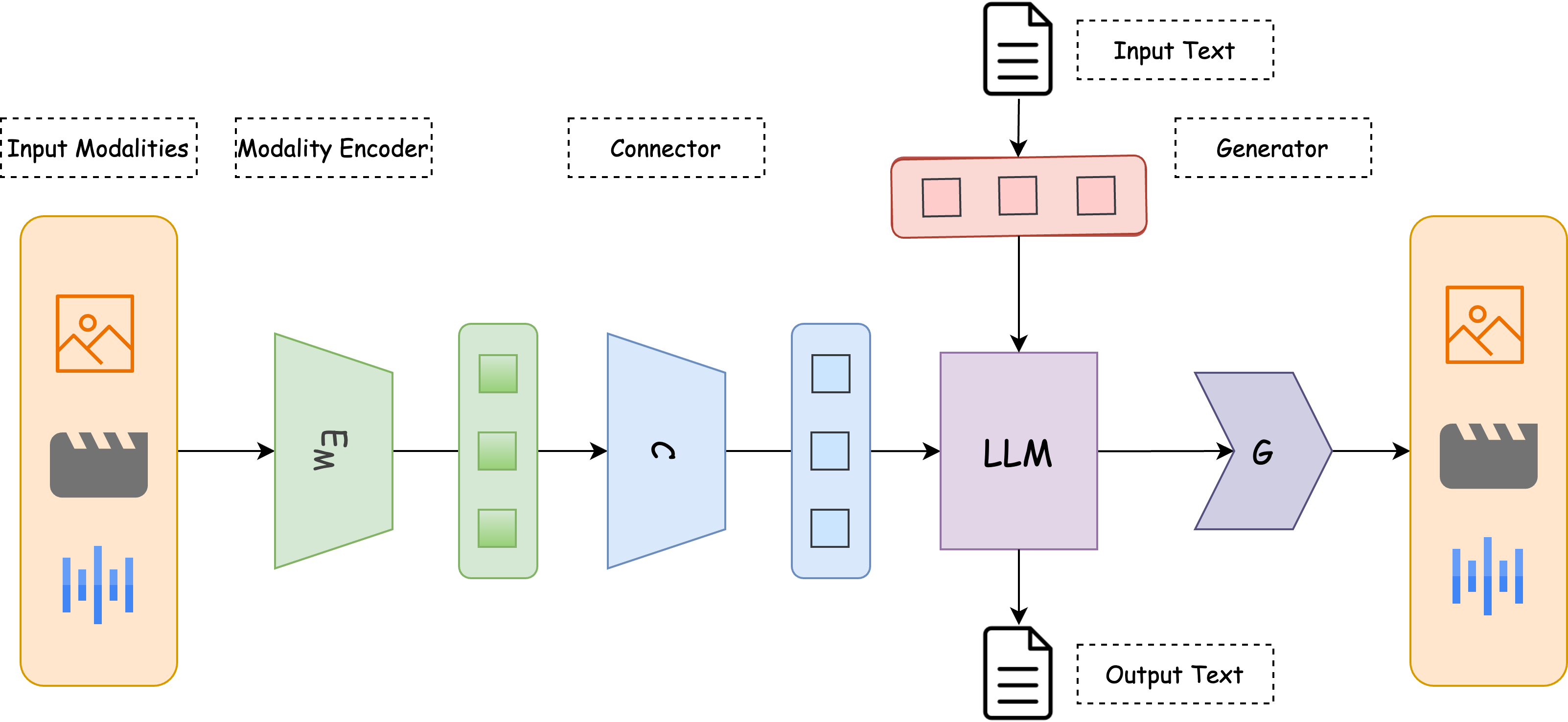}

  \caption{Schematic Representation of a Standard Multimodal Large Language Model (MLLM) Architecture}

  \label{fig:multimodal}

\end{figure*}

Multi-modal large language models (MLLMs) are designed to process and integrate information from multiple data sources, such as text, images, and audio, to perform a variety of tasks. These models leverage deep learning techniques to understand and generate content across different modalities, enhancing their applicability in real-world scenarios. For instance, Visual ChatGPT combines text and visual inputs to address complex queries \cite{wu2023visual}, while systems like BLIP-2 utilize a Qformer to integrate visual features with textual data for enhanced image-text interactions \cite{li2023blip}. MLLMs are particularly effective in tasks like visual question answering (VQA), where they can interpret and respond to queries based on visual content. The integration of modalities allows these models to offer more comprehensive responses and handle a broader range of interactions than single-modality models. The iterative training processes, often involving stages of freezing certain components while fine-tuning others, enable these models to maintain robust language capabilities while adapting to new modalities and tasks.

Figure \ref{fig:multimodal} displays a typical MLLM architecture, comprising an encoder $E_M$, a connector $C$, and a Large Language Model (LLM). Additionally, a generator $G$ can be integrated with the LLM to produce outputs beyond text, such as other modalities. The encoder processes inputs like images, audio, or videos into features, which the connector refines to enhance the LLM's comprehension capabilities. Connectors in these systems come in three main varieties: projection-based, query-based, and fusion-based. The first two types utilize token-level fusion, converting features into tokens that are combined with text tokens, whereas the fusion-based connector performs a feature-level fusion directly within the LLM \cite{yin2023survey}.

Recently, the integration of the Mixture of Experts (MoE) architecture into multi-modal large language models (MLLMs) has significantly advanced their capabilities. This approach employs multiple specialized sub-models, each fine-tuned for specific types of data or tasks such as image recognition or language processing. By selectively activating the most relevant experts based on the input and task, MoE allows MLLMs to dynamically adapt to the demands of multimodal data integration. This enhances the precision of the model in handling complex multimodal interactions and optimizes computational resources. Models like MoVA \cite{zong2024mova} and MoE-LLaVA \cite{lin2024moe} leverage MoE strategies effectively, improving performance while maintaining manageable computational costs during both training and inference phases. The adaptability and efficiency of MoE within MLLMs thus contribute significantly to their scalability and efficacy in real-world applications across varied tasks and data types \cite{li2024cumo}.

\section{Large Language Models in Healthcare and Medical Domain}
\label{sec:MedicalApp}

Language models have become a revolutionary force in the constantly changing world of healthcare and medicine, revolutionising how medical researchers and practitioners engage with data, patients, and huge corpus of medical knowledge \cite{thirunavukarasu2023large}. The use of language models in the medical field has undergone a significant metamorphosis, from the early days of simple rule-based systems, feature extraction, and keyword matching to the arrival of cutting-edge technologies like Transformers, and Large Language Models (LLMs) such as GPT-v4 \cite{openai2023gpt4}. These language models have overcome the constraints of conventional methods, enabling more complex natural language generation and interpretation.

Several pioneering large language models have significantly influenced the landscape of NLP. The emergence of the Transformer architecture \cite{vaswani2017attention} marked a significant milestone in the realm of natural language processing, leading to the emergence of expansive pre-trained language models like the BERT \cite{devlin2018bert} and RoBERTa \cite{liu2019roberta}.

BERT (Bidirectional Encoder Representations from Transformers), introduced by Devlin et al. (2018) \cite{devlin2018bert}, revolutionized NLP by pre-training a deep bidirectional model on a large corpus and outperforming previous models on various tasks. RoBERTa (A Robustly Optimized BERT Pretraining Approach) by Liu et al. (2019) \cite{liu2019roberta} demonstrated that further pre-training improvements and optimization could significantly enhance the performance of BERT.

In this section, we will first talk about the current large language models specifically for medical applications, in section~\ref{sec:MedModels}. Then, in section~\ref{sec:cases} we will talk about the use cases of various LLMs that designed mainly for patients, experts, and medical materials. 

\subsection{Large Language Models for Medical and Healthcare Applications}
\label{sec:MedModels}

Figure \ref{fig:scale_llm} provides a comprehensive overview of the progression in biomedical language model (LM) development from 2019 to 2023, emphasizing a logarithmic growth in model complexity and parameter count. It describes the evolutionary trajectories of various domain-specific adaptations of prominent models such as BioBERT, and GPT-2, along with the inception of more advanced systems like MedPaLM. The sizes of the illustrated models are proportional to their parameter volumes, showcasing a consistent trend towards larger, more capable models. This is culminated in the emergence of Large Language Models (LLMs) by 2023, which signifies a pivotal shift towards architectures with substantially heightened computational requirements and potential performance in biomedical text analysis and generation tasks.

\begin{table*}[!t]
\centering
\caption{Summary of Large Language Models in the Healthcare Space}
\label{tab:papers_table}
\begin{tabular}{|p{2.2cm}|c|p{3.8cm}|p{3.5cm}|p{1.6cm}|}
\hline
Method           & Year & Task                                                                                    & Institution                                                & Source Code                                          \\ \hline

BioMistral \cite{labrak2024biomistral}       & 2024 & Medical Question Answering                                                              & Avignon Université, Nantes Université                                  &  \href{https://huggingface.co/BioMistral/BioMistral-7B}{model}                                                    \\ \hline

Med-PaLM 2 \cite{singhal2023towards}       & 2023 & Medical Question Answering                                                              & Google Research, DeepMind                                  &                                                      \\ \hline
Radiology-Llama2 \cite{liu2023radiology} & 2023 & Radiology                                                                               & University of Georgia                                      &                                                      \\ \hline
DeID-GPT \cite{liu2023deid}         & 2023 & De-identification                                                                       & University of Georgia                                      & \href{https://github.com/yhydhx/ChatGPT-API}{code}                \\ \hline
Med-HALT \cite{umapathi2023med}    & 2023 & Hallucination test                                                                      & Saama AI Research  & \href{https://github.com/medhalt/medhalt}{code}                  \\ \hline                        
ChatCAD \cite{zhao2023chatcad+}         & 2023 & Computer-aided diagnosis                                                                & ShanghaiTech University                                    & \href{https://github.com/zhaozh10/ChatCAD}{code}                  \\ \hline
BioGPT \cite{luo2022biogpt}        & 2023 & Classification, relation extraction, question answering, etc.                           & Microsoft Research                                         & \href{https://github.com/microsoft/BioGPT}{code}                  \\ \hline
GatorTron \cite{yang2022gatortron}        & 2022 & Semantic textual similarity, natural language inference, and medical question answering & University of Florida                                               & \href{https://github.com/uf-hobi-informatics-lab/GatorTron}{code} \\ \hline
BioMedLM         & 2022 & Biomedical question answering                                                           & Stanford CRFM, MosaicML                                       & \href{https://github.com/stanford-crfm/BioMedLM}{code}            \\ \hline
BioBART \cite{yuan2022biobart}        & 2022 & Dialogue, summarization, entity linking, and NER                                        & Tsinghua University, International Digital Economy Academy & \href{https://github.com/GanjinZero/BioBART}{code}                \\ \hline
ClinicalT5 \cite{lu2022clinicalt5}     & 2022 & Classification, NER                                                                     & University of Oregon, Baidu Research                        & \href{https://huggingface.co/xyla/Clinical-T5-Large}{model}        \\ \hline
KeBioLM \cite{yuan2021improving}       & 2021 & Biomedical pre-training, NER, and relation extraction                                   & Tsinghua University, Alibaba Group                         & \href{https://github.com/GanjinZero/KeBioLM}{code}                \\ \hline

CRNN \cite{raj2017learning}       & 2017 & Relation classification                               & Indian Institute of Technology                         & \href{https://github.com/desh2608/crnn-relation-classification}{code}                \\ \hline

LSTM RNN \cite{lyu2017long}       & 2017 & Named entity recognition                               & Wuhan University                         & \href{https://github.com/lvchen1989/BNER}{code}                \\ \hline

\end{tabular}
\end{table*}

On the other hand, table \ref{tab:papers_table} provides an insightful overview of leading large language models within the healthcare domain. Recently, "BioMistral" was published as a a collection of open-source pre-trained large language
models for medical domains. In 2023, "Med-PaLM 2" and "Radiology-Llama2" emerged as key players, addressing medical question answering and radiology tasks, respectively. The "DeID-GPT" model extends its capabilities to de-identification, while "Med-HALT" specializes in hallucination testing. Simultaneously, "ChatCAD" offers valuable support in the realm of computer-aided diagnosis. "BioGPT" showcases versatility by handling classification, relation extraction, and question answering. "GatorTron" excels in semantic textual similarity and medical question answering, whereas "BioMedLM" narrows its focus to biomedical question answering. "BioBART" demonstrates prowess in dialogue, summarization, entity linking, and NER. "ClinicalT5" tackles classification and NER, while "KeBioLM" specializes in biomedical pre-training, NER, and relation extraction. Before the advent of language models or transformers, convolutional and recurrent neural networks represented the state of the art in the field. These models collectively represent remarkable strides in healthcare NLP, providing accessible source code or models for further exploration and practical application.

\subsection{Use Cases of Large Language Models in Healthcare}
\label{sec:cases}
In recent years, the emergence of large language models has catalyzed a transformative shift in the healthcare landscape, offering unprecedented opportunities for innovation and advancement. The capability of comprehending and generating text that resembles that of humans has demonstrated remarkable potential across a wide range of healthcare applications \cite{dasgupta2022language}. The applications of large language models in the healthcare sector are experiencing rapid growth. These models are being utilized for clinical decision support, medical record analysis, patient engagement, health information dissemination, etc. Their implementation holds the prospect to improve diagnostic accuracy, streamline administrative procedures, and ultimately enhance the efficiency, personalization, and comprehensiveness of healthcare delivery. This section delves into a comprehensive exploration of the multifaceted applications of large language models in healthcare, shedding light on their profound implications these applications bear on the trajectory of medical practices and the eventual outcomes experienced by patients.

\begin{figure*}

 \center

  \includegraphics[width=0.80\textwidth]{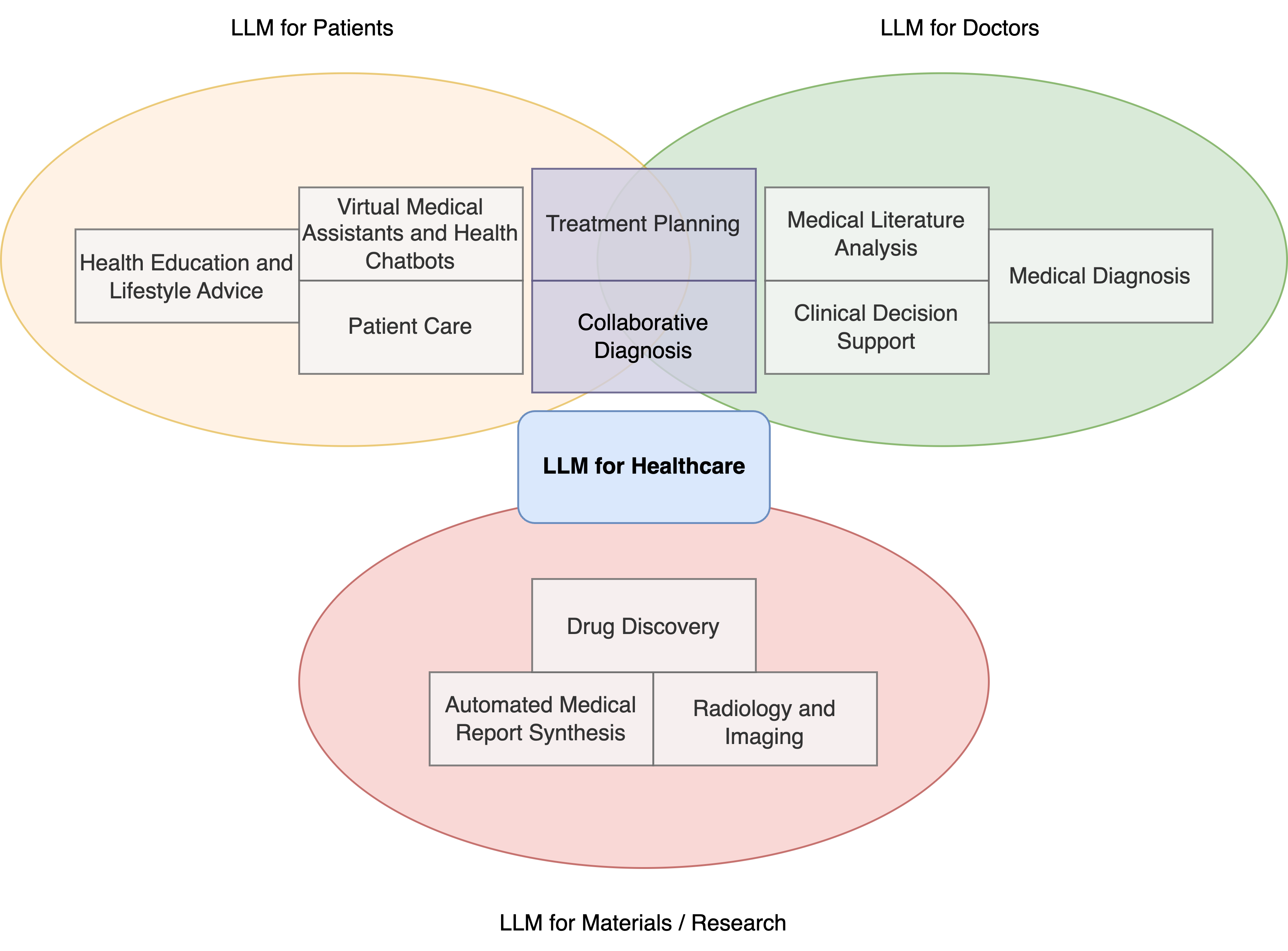}

  \caption{Applications of Large Language Models in Healthcare}

  \label{model_size}

\end{figure*}

\begin{itemize}
\item \textbf{Medical Diagnosis:} Certain clinical procedures may depend on the use of data analysis, clinical research, and recommendations \cite{singhal2023large, chen2023language}. LLMs may potentially contribute to medical diagnosis by conducting analyses on patient symptoms, medical records, and pertinent data, potentially aiding in the identification of potential illnesses or conditions with a certain degree of accuracy. Large language models have the potential to contribute to several aspects such as clinical decision assistance, clinical trial recruiting, clinical data administration, research support, patient education, and other related areas \cite{xue2023potential, chen2023boosting}. Corroborating this perspective, authors introduce a methodology that utilizes transformer models, namely BERT, RoBERTa, and DistilBERT, for the purpose of predicting COVID-19 diagnosis based on textual descriptions of acute alterations in chemosensation \cite{li2023text}. Similarly, a number of alternative investigations have been undertaken within the literature, proposing strategies using large language models for the diagnosis of Alzheimer's disease \cite{mao2023ad} and dementia \cite{agbavor2022predicting}. Furthermore, a corpus of literature has emerged, advocating the integration of large language model chatbots to cater to analogous objective \cite{bill2023fine, balas2023conversational, lai2023psy, bilal2023enhancing}.
   \item \textbf{Patient Care:} Large Language Models have emerged as transformative tools with the capacity to significantly enhance the realm of patient care \cite{javaid2023chatgpt}. Through the provision of personalised recommendations \cite{ali2023using}, customised treatment strategies, and continual monitoring of patients' advancements throughout their medical journeys \cite{nguyen2023application}, LLMs offer the promise of revolutionizing healthcare delivery. By harnessing the capabilities of LLMs, healthcare providers can ensure a more personalized and patient-centric approach to care. This technology enables the delivery of precise and well-informed medical guidance \cite{walker2023reliability}, aligning interventions with patients' distinct requirements and circumstances.

 The effective use of LLMs within clinical practise not only enhances patient outcomes but also enables healthcare professionals to make data-driven decisions, leading to enhanced patient care. As LLMs continue to advance, the potential for augmenting patient care through personalized recommendations and ongoing monitoring remains a promising trajectory in modern medicine \cite{iftikhar2023docgpt}. In essence, LLMs represent a pivotal leap forward, holding the capacity to reshape the landscape of patient care by fostering precision, adaptability, and patient-centeredness \cite{yang2023exploring}.

   \item \textbf{Clinical Decision Support}: Language models (LMs) have evolved into crucial decision support tools for healthcare professionals. By analyzing extensive medical data, LMs can provide evidence-based recommendations, enhancing diagnostic accuracy, treatment selection, and overall patient care. This fusion of artificial intelligence with healthcare expertise holds immense promise for improved medical decision-making. A body of existing research has illuminated promising prospects for the application of language models within clinical decision support, particularly within the domains of radiology \cite{wang2023chatcad}, oncology \cite{sorin2023large} and dermatology \cite{matin2023leveraging}.

   \item  \textbf{Medical Literature Analysis}: Large language models (LLMs) exhibit remarkable efficiency in comprehensively reviewing and succinctly summarizing extensive volumes of medical literature. This capability aids both researchers and clinicians in maintaining topicality with cutting-edge developments and evidence-based methodologies, ultimately fostering informed and optimized healthcare practices. In a fast-evolving field like healthcare, the ability to maintain currency with the latest advancements is paramount, and LLMs can play a pivotal role in ensuring that healthcare remains at the forefront of innovation and evidence-based care delivery \cite{sallam2023utility, tang2023evaluating}.

   \item \textbf{Drug Discovery}: Large Language Models, have a significant impact in facilitating drug discovery through their capacity to scrutinize intricate molecular structures, discern promising compounds with therapeutic potential, and forecast the efficacy and safety profiles of these candidates \cite{liu2021ai, datta2022bert}. Chemical language models have exhibited notable achievements in the domain of de novo drug design \cite{grisoni2023chemical}. 
In this corresponding study, authors explore the utilization of pre-trained biochemical language models to initialize targeted molecule generation models, comparing one-stage and two-stage warm start strategies, as well as evaluating compound generation using beam search and sampling, ultimately demonstrating that warm-started models outperform baseline models and the one-stage strategy exhibits superior generalization in terms of docking evaluation and benchmark metrics, while beam search proves more effective than sampling for assessing compound quality \cite{uludougan2022exploiting}.

    \item \textbf{Virtual Medical Assistants and Health Chatbots}: LLMs may also serve as the underlying intelligence for health chatbots, revolutionizing the healthcare landscape by delivering continuous and personalized health-related support. These chatbots can offer medical advice, monitor health conditions, and even extend their services to encompass mental health support, a particularly pertinent aspect of healthcare given the growing awareness of mental well-being \cite{bilal2023enhancing, bill2023fine}.

    \item \textbf{Radiology and Imaging:} Multi-modal visual-language models, through their integration of visual and textual data, hold significant promise for augmenting medical imaging analysis. Radiologists can benefit from these models as they facilitate the early identification of abnormalities in medical images and contribute to the generation of more precise and comprehensive diagnostic interpretations, ultimately advancing the accuracy and efficiency of diagnostic processes in the field of medical imaging \cite{ma2023cephgpt, wang2023chatcad, khader2022medical, thawkar2023xraygpt, liu2023chatgpt, monajatipoor2022berthop, roshanzamir2021transformer}.

\item \textbf{Automated Medical Report Synthesis from Imaging Data:} Automated medical report generation from images is crucial for streamlining the time-consuming and error-prone task faced by pathologists and radiologists. This emerging field at the intersection of healthcare and artificial intelligence (AI) aims to alleviate the burden on experienced medical practitioners
and enhance the accuracy of less-experienced ones. The integration of AI with medical imaging facilitates the automatic drafting of reports, encompassing abnormal findings, relevant normal observations, and patient history. Early efforts employed data-driven neural networks, combining convolutional and recurrent models for single-sentence reports, but limitations arose in capturing the complexity of real medical scenarios \cite{he2023survey}. Recent advances leverage large language models (LLMs) such as ChatCAD \cite{wang2023chatcad}, enabling more sophisticated applications. ChatCAD enhances medical-image Computer-Aided Diagnosis networks, yielding significant improvements in report generation. ChatCAD+ further addresses writing style mismatches, ensuring universality and reliability across diverse medical domains, incorporating a template retrieval system for consistency with human expertise \cite{zhao2023chatcad+}. In \cite{giorgi2023wanglab}, authors use pre-trained language model (PLM) and in-context learning (ICL) to generate clinical note from doctor patient conversation. These integrated systems signify a pivotal advancement in automating medical report generation through the strategic utilization of LLMs.

\end{itemize}

\subsection{Explainable AI Methods for Interpreting Healthcare LLMs}

Large Language Models (LLMs) have significantly advanced the healthcare domain, enhancing tasks such as medical diagnosis and patient monitoring. However, the complexity of these models necessitates interpretability for reliable decision-making \cite{huang2024explainable}. This section discusses "eXplainable and Interpretable Artificial Intelligence" (XIAI) and examines recent XIAI methods by their functionality and scope. Despite challenges, such as the difficulty in quantifying interpretability and the lack of standardized evaluation metrics, opportunities exist in integrating XIAI to add interpretability for LLMs in healthcare. Notable XIAI methods include SHAP \cite{thorsen2022discrete}, which quantifies feature contributions, LIME \cite{zhang2019explainable, ozyegen2022word}, which generates interpretable models through input perturbations, t-SNE for visualizing high-dimensional data \cite{dobrakowski2021interpretable}, attention mechanisms that highlight key features \cite{huang2019clinicalbert}, and knowledge graphs that structure contextual relationships \cite{gao2023leveraging}, all of which provide crucial insights into model decision-making processes.

Existing research delves into explainability for LLMs in the healthcare domain. For instance, Yang et al. (2023) \cite{yang2023towards} investigate different prompting strategies using emotional cues and expert-written examples for mental health analysis with LLMs. This study shows that models like ChatGPT can generate near-human-level explanations, enhancing interpretability and performance. Additionally, ArgMedAgents (Hong et al., 2024) \cite{hong2024argmed} is a multi-agent framework designed for explainable clinical decision reasoning through interaction, utilizing the Argumentation Scheme for Clinical Discussion and a symbolic solver to provide clear decision explanations. Furthermore, Gao et al. (2023) propose enhancing LLM explainability for automated diagnosis by integrating a medical knowledge graph (KG) from the Unified Medical Language System (UMLS), using the DR.KNOWS model to interpret complex medical concepts. Their experiments with real-world hospital data demonstrate a transparent diagnostic pathway. Similarly, TraP-VQA \cite{gao2023leveraging}, a novel vision-language transformer for Pathology Visual Question Answering (PathVQA), employs Grad-CAM and SHAP methods to offer visual and textual explanations, ensuring transparency and fostering user trust.

We have compiled a list in table \ref{tab:xai_table}, detailing XIAI attributes, summarizing recent research works focused on explainability methods for LLMs in the healthcare domain. This table includes evaluations of various models, highlighting their unique contributions to enhancing interpretability and reliability in medical applications. Each entry outlines the task, method, XAI attributes, and evaluation metrics, offering a clear overview of the advancements and effectiveness of XIAI techniques in improving decision-making processes in healthcare.

\begin{table*}[!t]
\centering
\caption{Summary of Recent XIAI Methods for LLMs in Healthcare}
\label{tab:xai_table}
\resizebox{\textwidth}{!}{
\begin{tabular}{|p{2.2cm}|c|p{4.5cm}|p{5.5cm}|p{3.0cm}|}
\hline
\textbf{Method} & \textbf{Year} & \textbf{Task} & \textbf{XIAI Attributes} & \textbf{XIAI Evaluation Metric} \\ \hline
MentaLLaMA [\href{https://github.com/SteveKGYang/MentalLLaMA}{code}] \cite{yang2024mentallama} & 2024 & Mental health analysis & Prompt-based (ChatGPT w/ task-specific instructions) & BART-score, Human Eval \\ \hline
ArgMed-Agents \cite{hong2024argmed} & 2024 & Clinical decision reasoning & Prompt-based (Self-argumentation iterations + symbolic solver) & Pred. accuracy with LLM evaluator \\ \hline
Diagnostic reasoning prompts \cite{savage2024diagnostic} & 2024 & Medical Question Answering (MedQA) & Prompt-based (Bayesian, differential diagnosis, analytical, and intuitive reasoning) & Expert Evaluation, Inter-rater agreement \\ \hline
SkinGEN \cite{lin2024skingen} & 2024 & Dermatological diagnosis & Visual explanations (Stable Diffusion), interactive framework & Perceived explainability ratings \\ \hline
DR. KNOWS \cite{gao2023leveraging} & 2023 & Automated diagnosis generation & Knowledge Graph (explainable diagnostic pathway) & - \\ \hline
Human-AI Collaboration \cite{lee2023understanding} & 2023 & Clinical decision making & Salient features, counterfactual explanations & Agreement Level, Usability Questionnaires \\ \hline
ChatGPT \cite{yang2023towards} & 2023 & Mental health analysis & Prompt-based (emotional cues and expert-written few-shot examples) & BART-score, Human Eval \\ \hline
CHiLL \cite{mcinerney2023chill} & 2023 & Clinical predictive tasks, Chest X-ray report classification & Interpretable features, linear models & Expert Evaluation, Clinical Judgement Alignment \\ \hline
Trap-VQA \cite{naseem2022vision} & 2022 & Pathology Visual Question Answering (PathVQA) & Grad-CAM, SHapley Additive exPlanations & Qualitative Evaluation \\ \hline
Vision Transformer \cite{park2103vision} & 2021 & Covid-19 diagnosis & Saliency maps & Visualisation \\ \hline
ClinicalBERT [\href{https://github.com/kexinhuang12345/clinicalBERT}{code}] \cite{huang2019clinicalbert} & 2019 & Predicting hospital readmission & Attention weights & Visualisation \\ \hline
\end{tabular}
}
\end{table*}

\subsection{Future Trajectories of Large Language Models in Healthcare}

As large language models (LLMs) continue to integrate into the healthcare sector, future developments promise to revolutionize patient care and medical research. A particularly promising avenue involves enhancing LLMs' capabilities to interpret and generate not only textual but also biomolecular data \cite{pan2023large}. This advancement could significantly improve applications in genomics and personalized medicine, enabling these models to predict individual responses to treatments based on genetic profiles, thereby advancing the precision of medical interventions. Furthermore, incorporating adaptive learning capabilities in real-time could transform LLMs into dynamic aids during surgical procedures or emergencies, where they might analyze data from medical devices on-the-fly \cite{liang2024online} to offer critical decision support.

Another innovative trajectory for LLMs in healthcare is the development of federated learning systems \cite{che2023federated}. Such systems could facilitate the secure, privacy-preserving propagation of medical knowledge across institutions, improving model robustness and applicability across varied demographic groups without direct data sharing. This approach will not only enhance the privacy and security of patient data but will also enable a collective intelligence that could lead to more generalized healthcare solutions.

The potential of large language models (LLMs) in healthcare extends into the realms of explainable medical AI \cite{zhao2024explainability} and the utilization of multi-modal models incorporating sensor data. By integrating LLMs with wearable technologies \cite{kim2024health}, these advanced models can serve as continuous health monitors in non-clinical settings. 

To further advance explainable medical AI, LLMs can be instrumental in deciphering the complexities of medical conditions and treatment outcomes. By processing and interpreting multi-modal data, including sensor readings, these models can contribute to a deeper understanding of patient health on a granular level. This may aid in the development of precise, targeted therapies, improving patient outcomes and enhancing the transparency of medical decisions. 

Large Language Models (LLMs) are poised to revolutionize the healthcare domain by enhancing diagnostic accuracy, personalizing treatment plans, and optimizing operational efficiencies. By integrating LLMs into electronic health record systems, healthcare providers can more accurately diagnose conditions through natural language processing techniques that analyze clinical notes and patient histories. Moreover, LLMs assist in generating personalized treatment recommendations by analyzing vast datasets that include genetic information, clinical outcomes, and patient preferences. Furthermore, these models streamline administrative tasks by automating documentation, coding, and billing processes, thus reducing operational costs and allowing medical staff to focus more on patient care. As generative AI advances, its transformative impact on the healthcare sector is becoming increasingly significant. This technology is poised to revolutionize areas such as clinical trials, personalized medicine, and drug discovery. Additionally, its applications extend to enhancing natural language processing and understanding, improving medical imaging, and supporting virtual assistants in patient care. Generative AI also plays a crucial role in illness detection and screening, facilitating more accurate diagnostics. Moreover, it is being integrated into medical conversation tasks, voice generation, video generation, and image synthesis and manipulation within healthcare settings \cite{pahune2024large}. These innovations are not only improving the efficiency of medical services but are also paving the way for new methods of patient interaction and treatment planning. As these applications continue to mature, LLMs will become integral in transforming healthcare services into more efficient, accurate, and personalized systems.

\begin{figure*}

 \center

  \includegraphics[width=0.80\textwidth]{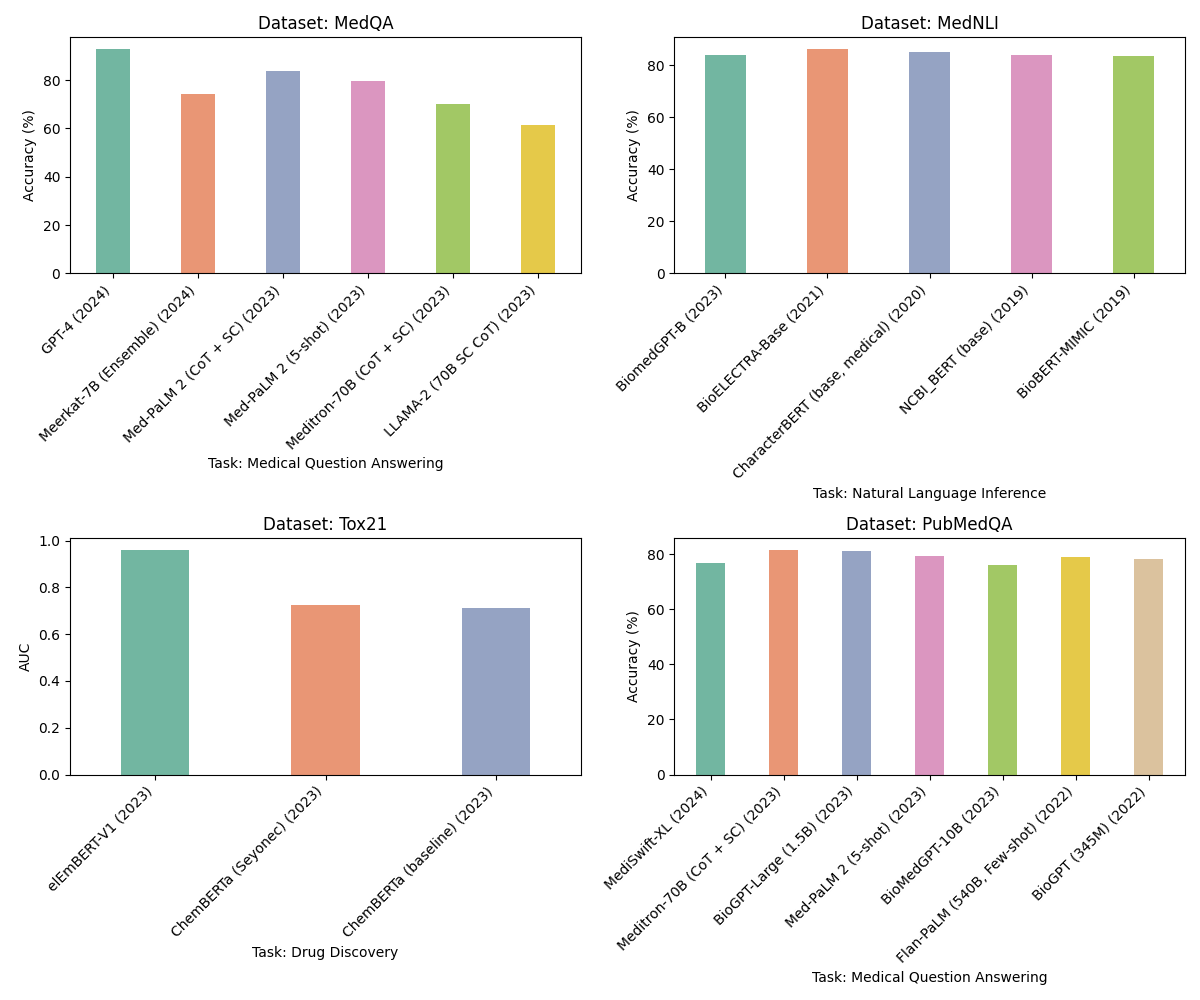}

  \caption{Comparative Performance of Healthcare LLMs}

  \label{fig:medllm_comp}

\end{figure*}

\subsection{Performance Evaluation and Benchmarks}

The medicine and healthcare industries largely acknowledge the potential of artificial intelligence (AI) to drive substantial progress in the delivery of healthcare. However, empirical evaluations have demonstrated that numerous artificial intelligence (AI) systems do not successfully achieve their desired translation goals, primarily because of intrinsic deficiencies that become evident only after implementation \cite{reddy2021evaluation, reddy2023evaluating}. In order to optimize the utilization of language models (LLMs) within healthcare settings, it is imperative to develop evaluation frameworks that possess the capacity to thoroughly evaluate their safety and quality. It is important to note that certain highly effective models, such as ChatGPT and PaLM 2 \cite{anil2023palm}, are now not publicly available. The absence of accessibility gives rise to notable problems pertaining to transparency, which is a crucial factor in the medical domain and hinders the capacity to thoroughly examine the structure and results of the model. Consequently, this impedes endeavors to recognize and address biases and hallucinations. Thorough research is necessary to understand the specific performance characteristics and ramifications of utilizing publicly accessible, pre-trained language models in addressing the challenges in the healthcare and medical domains. Language models that have been pre-trained using medical data also encounter comparable difficulties. Therefore, the careful choice and implementation of suitable performance metrics to evaluate the language model assume great significance.

In table \ref{tab:eval_metrics}, we present a comprehensive catalog of performance metrics, including but not limited to the F1 score, BLEU, GLUE, and ROGUE, which constitute the standard evaluative criteria employed for the rigorous assessment of large language models operating within the healthcare and medical domain. This compendium of metrics serves as a valuable reference, encapsulating the quantitative and qualitative measures utilized to gauge the efficacy, proficiency, and suitability of these models in diverse healthcare applications \cite{reddy2023evaluating}.

\begin{table*}[!t]
\centering
\caption{LLM Performance Benchmark}
\label{tab:llm_perf_bench}
\resizebox{\textwidth}{!}{
\begin{tabular}{|p{2cm}|c|p{4cm}|p{5cm}|c|}
\hline
\textbf{Organization} & \textbf{Model} & \textbf{MMLU Score} & \textbf{Coding (HumanEval)} & \textbf{Release Date} \\ \hline
OpenAI & GPT-4 Opus & 88.7 & - & May 2024 \\ \hline
Anthropic & Claude 3.5 Sonnet & 88.7 & 92.0 & June 2024 \\ \hline
Anthropic & Claude 3 Opus & 86.8 & - & March 2024 \\ \hline
OpenAI & GPT-4 Turbo & 86.4 & 85.4 & April 2024 \\ \hline
OpenAI & GPT-4 & 86.4 & 90.2 & April 2023 \\ \hline
Meta & Llama 3 400B & 86.1 & - & - \\ \hline
Google & Gemini 1.5 Pro & 85.9 & 84.1 & May 2024 \\ \hline
Google & Gemini Ultra & 83.7 & - & December 2023 \\ \hline
OpenAI & GPT-3.5 Turbo & - & 73.2 & - \\ \hline
Meta & Llama 3 (70B) & - & 81.7 & - \\ \hline
Meta & Llama 3 (8B) & - & 62.2 & - \\ \hline
Google & Gemini 1.5 Flash & - & 74.3 & - \\ \hline
\end{tabular}
}
\end{table*}

\subsection{Quantitative Performance Comparison of LLMs in Healthcare Domain}

Recent advancements in language models have been benchmarked against diverse datasets to evaluate their capabilities across various domains. One such comprehensive benchmark is the MMLU (Massive Multitask Language Understanding) \cite{hendrycks2020measuring}, designed to assess the understanding and problem-solving abilities of language models. The MMLU comprises 57 tasks spanning topics such as elementary mathematics, US history, computer science, and law, requiring models to demonstrate a broad knowledge base and problem-solving skills. This benchmark provides a standardized method to test and compare various language models, including OpenAI GPT-4o, Mistral 7b, Google Gemini, and Anthropic Claude 3, among others.

The HumanEval benchmark is used to measure the functional correctness of code generated by LLMs from docstrings. This benchmark evaluates models based on their ability to generate code that passes provided unit tests, using the pass@k metric. If any of the 'k' solutions generated by the model pass all unit tests, the model is considered successful in solving the problem \cite{chen2021evaluating}. Table \ref{tab:llm_perf_bench} provides a concise summary of the performance of various LLMs on the MMLU and HumanEval (Coding) datasets \cite{mmlu_benchmark}.

In the healthcare domain, a variety of LLMs have been developed and evaluated on specific datasets such as MedQA, MedNLI \cite{jin2019probing}, Tox21 \cite{mayr2016deeptox}, and PubMedQA \cite{jin2019pubmedqa}. The GPT-4 (2024) model stands out in the MedQA dataset with an impressive accuracy of 93.06\%, significantly outperforming other models like Med-PaLM 2 (CoT + SC) (2023), which achieves 83.7\%, and Meerkat-7B (Ensemble) (2024), with 74.3\%. In the MedNLI dataset, BioELECTRA-Base (2021) achieves the highest accuracy of 86.34\%, closely followed by CharacterBERT (base, medical) (2020) at 84.95\%. The Tox21 dataset highlights elEmBERT-V1 (2023) with an outstanding AUC of 0.961, making it the most effective in predicting chemical properties and toxicity. For the PubMedQA dataset, Meditron-70B (CoT + SC) (2023) and BioGPT-Large (1.5B) (2023) exhibit strong performance with accuracies of 81.6\% and 81.0\%, respectively \cite{medical_papers_with_code}. These findings underscore the variability in performance across different healthcare tasks, emphasizing the need for careful selection of models based on specific application requirements \cite{lee2023drug}. Figure \ref{fig:medllm_comp} presents a comparative performance analysis of various healthcare LLMs, highlighting their accuracy and AUC metrics across different datasets including MedQA, MedNLI, Tox21, and PubMedQA.

\begin{table*}[htbp]
\centering
\caption{\label{tab:eval_metrics} Evaluation Metrics for Language Models in Healthcare Domain}
\renewcommand{\arraystretch}{1.5} 
\scriptsize 
\begin{tabular}{|p{2cm}|p{4cm}|p{2cm}|p{3cm}|}
\hline
\textbf{Eval. Metric} & \textbf{Description} & \textbf{References} & \textbf{Key Highlights} \\
\hline
\multirow{4}{*}{Perplexity} & Perplexity, a probabilistic metric, quantifies the uncertainty in the predictions of a language model. Lower values indicate higher prediction accuracy and coherence. & \cite{liao2023differentiate} & - \\ \cline{3-4}
& & \cite{manoel2023federated} & The federated learning model achieved a best perplexity value of \textbf{3.41} for English. \\ \cline{3-4}
& & \cite{zhang2019vettag} & The Transformer model achieved a test perplexity of \textbf{15.6} on the PSVG dataset, significantly outperforming the LSTM's perplexity of \textbf{20.7}. \\ \cline{3-4}
& & \cite{gao2023leveraging} & The lowest perplexity achieved was \textbf{3.86e-13} with manually designed prompts. \\ \hline
\multirow{2}{*}{BLEU} & The BLEU score assesses the quality of machine translation by comparing it to reference translations. & \cite{wang2023clinicalgpt} & The best BLEU-1 score achieved was \textbf{13.9} by the ClinicalGPT model. \\ \cline{3-4}
& & \cite{li2023huatuo} & T-5 (fine-tuned) model achieved the best BLEU-1 score of \textbf{26.63}. \\ \hline
\multirow{2}{*}{GLEU} & GLEU score computes mean scores of various n-grams to assess text generation quality. & \cite{wang2023clinicalgpt} & The best GLEU score achieved was \textbf{2.2} by the Bloom-7B model. \\ \cline{3-4}
& & \cite{li2023huatuo} & T-5 (fine-tuned) model achieved the best GLEU score of \textbf{11.38}. \\ \hline
\multirow{2}{*}{ROUGE} & ROUGE score evaluates summarization and translation by measuring overlap with reference summaries. & \cite{wang2023clinicalgpt} & The best ROGUE-L score achieved was \textbf{21.3} by the ClinicalGPT model. \\ \cline{3-4}
& & \cite{li2023huatuo} & T-5 (fine-tuned) model achieved the best ROGUE-L score of \textbf{24.85}. \\ \hline
Distinct n-grams & Measures the diversity of generated responses by counting unique n-grams. & \cite{li2023huatuo} & On the Huatuo-26M dataset, the fine-tuned T5 model achieved Distinct-1 and Distinct-2 scores of \textbf{0.51} and \textbf{0.68}, respectively. \\ \hline
\multirow{4}{*}{F1 Score} & The F1 score balances precision and recall, measuring a model's accuracy in identifying positive instances and minimizing false results. & \cite{yang2022large} & The GatorTron-large model achieved the best F1 score of \textbf{0.9627} for medical relation extraction. \\ \cline{3-4}
& & \cite{yang2022gatortron} & The GatorTron-large model achieved the best F1 score of \textbf{0.9000} for clinical concept extraction and \textbf{0.9627} for medical relation extraction. \\ \cline{3-4}
& & \cite{crema2023advancing} & The multicenter Transformers-based model achieved an overall F1 score of \textbf{84.77\%} on the PsyNIT dataset. \\ \cline{3-4}
& & \cite{datta2022bert} & The BERT-D2 model achieved an F1 score of \textbf{81.97\%} on the DDI Extraction 2013 corpus. \\ \hline
\multirow{2}{*}{BERTScore} & BERTScore calculates similarity scores between tokens in candidate and reference sentences, using contextual embeddings. & \cite{zhang2019bertscore} & - \\ \cline{3-4}
& & \cite{giorgi2023wanglab} & The Longformer-Encoder-Decoder (LED\textsubscript{large-PubMed}) model achieved the best BERTScore F1 of \textbf{70.7}. \\ \hline
\multirow{1}{*}{Human Evaluation} & Involves expert human assessors rating the quality of model-generated content, providing qualitative insights into its performance. & \cite{beaulieu2023evaluating} & The median performance for all human SCORE users was \textbf{65\%}, whereas ChatGPT correctly answered \textbf{71\%} of multiple-choice SCORE questions and \textbf{68\%} of Data-B questions. \\ \cline{3-4}
\hline
\end{tabular}
\end{table*}

\section{Limitations and Open Challenges}
\label{sec:others}

The integration of large language models (LLMs) in healthcare presents complex challenges, including the need for explainability in model decision-making, robust security and privacy measures to protect sensitive patient data, addressing biases and ensuring fairness in medical AI applications, mitigating the issue of hallucinations where models generate erroneous information, and establishing clear legal frameworks for the responsible use of LLMs in healthcare, all of which demand careful scrutiny and resolution to harness the full potential of these models for improving healthcare outcomes while upholding ethical and legal standards.

\subsection{Model Explainability and Transparency}

Large language models face notable challenges when applied to healthcare. Their recommendations often lack transparency due to their opaque nature, which can hinder acceptance among healthcare professionals who prioritize explainability in medical decision-making. Moreover, the presence of biases in the training data may compromise the accuracy of these models, potentially leading to incorrect diagnoses or treatment recommendations. It is therefore crucial for medical professionals to exercise caution and thoroughly review and validate the recommendations provided by large language models before integrating them into their clinical decision-making processes \cite{ali2023chatgpt}. In healthcare, the importance of interpretability and explainability for AI models utilized in medical imaging analysis and clinical risk prediction cannot be overstated. Inadequate transparency and explainability have the potential to undermine trustworthiness and hinder the validation of clinical recommendations. Consequently, effective governance underscores the continuous pursuit of transparency and interpretable frameworks, aiming to augment the decision-making process in the realm of healthcare \cite{reddy2023evaluating}. Large language models (LLMs) often function as "blackboxes", rendering it challenging to discern the underlying processes leading to specific conclusions or suggestions. In the healthcare context, where the repercussions of decisions are profound, it becomes imperative for practitioners to grasp the logic behind AI-generated outputs. The persistent endeavor to create models that are more interpretable and transparent remains an enduring challenge within the healthcare domain \cite{briganti2023clinician, bisercic2023interpretable, jiang2023balanced}.

\begin{figure*}

 \center

  \includegraphics[width=0.80\textwidth]{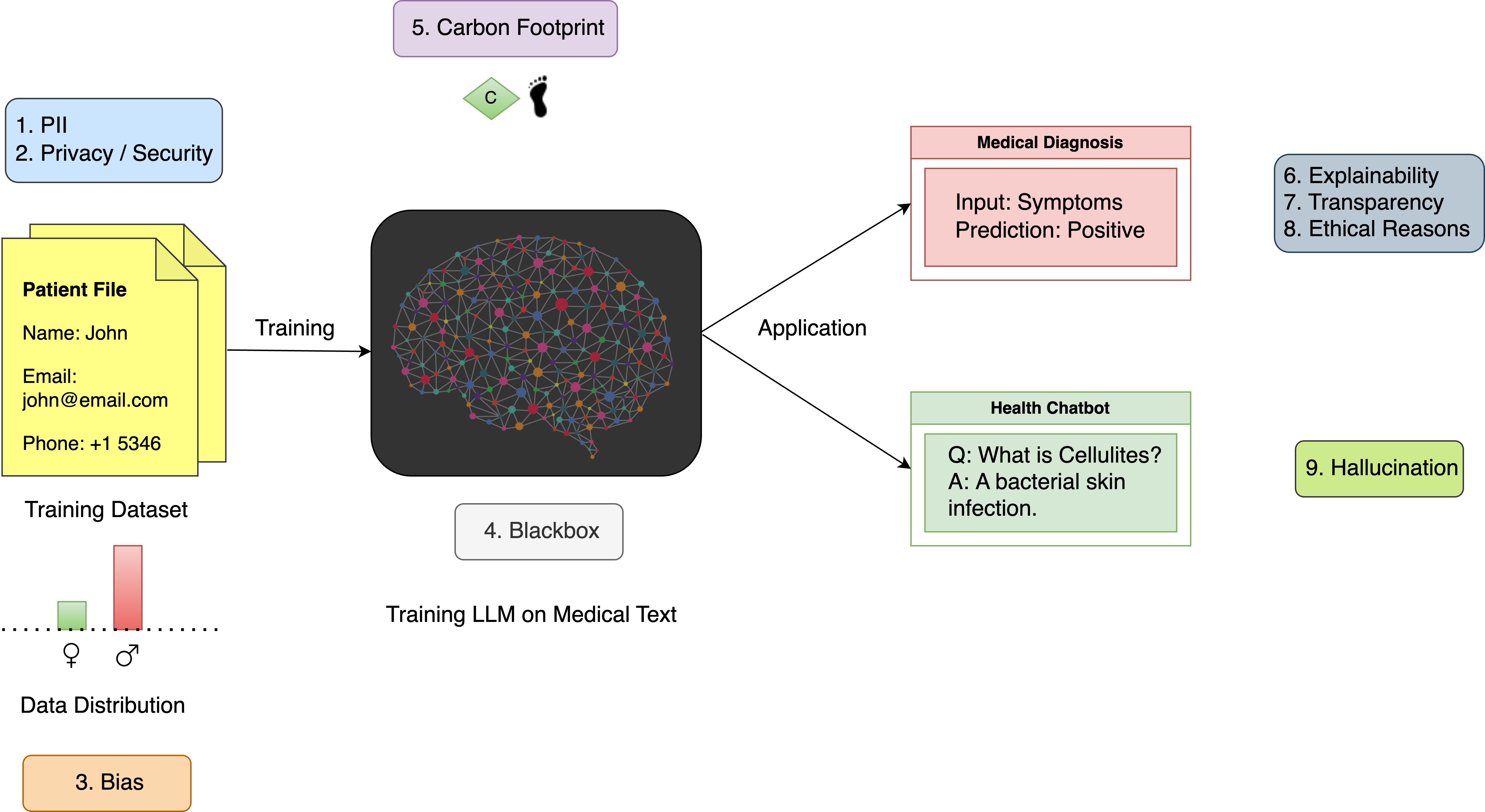}

  \caption{Challenges of Large Language Models in Healthcare}

  \label{model_size}

\end{figure*}

\subsection{Security and Privacy Considerations}
Large Language Models (LLMs) are used in medical research, which necessitates careful consideration of data privacy and security issues. Researchers are entrusted with the duty of managing extremely private patient data while enforcing rigorous compliance with current privacy laws. The use of LLMs in this setting raises concerns about a number of aspects of data processing, including as data protection, the possibility of re-identification, and the moral application of patient data. One notable issue is the inadvertent inclusion of personally identifiable information (PII) within pre-training datasets, which can compromise patient confidentiality. Additionally, LLMs can make privacy-invading inferences by deducing sensitive personal attributes from seemingly innocuous data, potentially violating individual privacy \cite{omiye2023large}. Implementing strong measures like data anonymization, safe data storage procedures, and steadfast adherence to ethical standards are essential to addressing these issues. Together, these steps make up crucial safeguards meant to protect research participants' trust, maintain the integrity of research processes, and protect patient privacy. The importance of these factors is underscored by the necessity of balancing the significant contributions of LLMs in medical research with the critical requirement to protect private patient information \cite{thapa2023chatgpt}. LLMs' ability to find potentially revealing patterns in large amounts of health data, even when anonymized, poses a serious privacy risk. This necessitates strict regulations and technical protections.  Anonymizing data more effectively is crucial, as are algorithms designed to spot and prevent the re-identification of individuals. Ongoing monitoring of what LLMs produce is vital to ensure privacy isn't accidentally compromised. Implementing these measures helps guarantee responsible use of sensitive data, allowing LLMs to be used ethically in healthcare while still respecting patient privacy. To ensure the ethical use of LLMs in healthcare, strong governance frameworks must extend beyond basic privacy laws.  Proactive policies should anticipate challenges, and experts need to verify LLMs meet ethical guidelines. Engaging patients and healthcare providers in the development process promotes transparency and maintains trust in how health data is used within these systems.

\subsection{Bias and Fairness}
Researching ways to tackle and reduce biases in language models, while also comprehending their ethical ramifications, represents a pivotal research domain. It is imperative to create techniques for identifying, alleviating, and forestalling biases in large language models. A primary concern associated with Large Language Models (LLMs) pertains to the risk of producing misinformation or biased outputs. These models, drawing from extensive text data, encompass both dependable and unreliable sources, which can inadvertently result in the generation of inaccurate or misleading information. Furthermore, if the training data incorporates biases, such as gender or racial biases prevalent within scientific literature, LLMs can perpetuate and magnify these biases in their generated content.

To ensure the reliability and accuracy of information derived from LLMs, researchers must exercise caution and implement rigorous validation and verification processes. LLMs have the potential to amplify pre-existing biases inherent in their training data, particularly those linked to demographics, disease prevalence, or treatment outcomes. Consequently, the generated outputs may inadvertently reflect and perpetuate these biases, posing considerable challenges in achieving equitable and unbiased healthcare outcomes.

To address these challenges, researchers must remain vigilant in recognizing and mitigating biases within both the training data and the outputs generated by LLMs. This diligence is crucial for promoting fairness and inclusivity within the realm of biomedical research and healthcare applications, ultimately enhancing the ethical and equitable utility of LLMs in these domains \cite{thapa2023chatgpt}. Prioritizing bias mitigation in LLMs is essential. Researchers should curate and preprocess training data diligently to reduce inherent biases and address sources of inequality. Routine audits and evaluations are necessary to identify and correct biases in model training and deployment. Collaborative efforts between domain experts, ethicists, and data scientists can establish guidelines and best practices for unbiased LLM development, fostering fairness and inclusivity in biomedical research and healthcare.

\subsection{Hallucinations and Fabricated Information}

Language models exhibit a proclivity for generating erroneous content, commonly referred to as hallucinations. This phenomenon is characterized by the production of text that appears plausible but lacks factual accuracy. This inherent trait poses a substantial risk when such generated content is employed for critical purposes, such as furnishing medical guidance or contributing to clinical decision-making processes. The consequences of relying on hallucinatory information in healthcare contexts can be profoundly detrimental, potentially leading to harmful or even catastrophic outcomes \cite{tian2023opportunities}.

The gravity of this issue is exacerbated by the continuous advancement of Large Language Models (LLMs), which continually enhance their capacity to generate increasingly persuasive and believable hallucinations. Moreover, LLMs are often critiqued for their opacity, as they provide no discernible link to the original source of information, thereby creating a formidable barrier to the verification of the content they produce. To mitigate these risks, healthcare professionals must exercise extreme caution when utilizing LLMs to inform their decision-making processes, rigorously validating the accuracy and reliability of the generated information.

Current research endeavors are dedicated to addressing hallucination issues within Large Language Models (LLMs) in the healthcare and medical domain. The introduction of Med-HALT, a novel benchmark dataset, serves the purpose of evaluating hallucination phenomena in LLMs in medical contexts. Med-HALT encompasses two distinct test categories: reasoning-based and memory-based hallucination assessments. These tests have been meticulously designed to gauge the problem-solving and information retrieval capabilities of LLMs when operating within the medical domain \cite{umapathi2023med}.

\subsection{Legal and Ethical Reasons}

Ethical concerns extend to the generation of potentially harmful content by LLMs, especially when delivering distressing medical diagnoses without providing adequate emotional support. Moreover, the blurring line between LLM-generated and human-written text poses a risk of misinformation dissemination, plagiarism, and impersonation.

To address these challenges, rigorous auditing and evaluation of LLMs are essential, along with the development of regulations for their medical use. Thoughtful selection of training datasets, particularly within the medical domain, is crucial to ensure the responsible handling of sensitive data. These measures collectively strive to strike a balance between harnessing LLMs' potential and safeguarding patient privacy and ethical standards \cite{omiye2023large}. 

The European Union’s AI Act and the United States’ Health Insurance Portability and Accountability Act (HIPAA) are two significant regulatory frameworks impacting the deployment of AI in healthcare. The AI Act introduces comprehensive regulations, including the Artificial Intelligence Liability Directive (AILD), which addresses liability for AI-related damages. This directive ensures that victims are compensated and that preventive measures are cost-effective. The AI Act classifies General Purpose AI (GPAI) models and imposes specific obligations on providers, including technical documentation, risk assessments, and transparency about training data \cite{novelli2024generative}.

In the United States, HIPAA sets stringent standards for the protection of patient data, impacting how LLMs handle sensitive information. Compliance with HIPAA requires robust data encryption, regular security assessments, and strict access controls to protect patient information. These regulations ensure that LLMs used in healthcare settings adhere to high standards of privacy and security, mitigating risks associated with data breaches and unauthorized access.

Other relevant laws and compliance frameworks include the General Data Protection Regulation (GDPR) in the EU, which emphasizes data protection and privacy, and the Medical Device Regulation (MDR) that ensures the safety and efficacy of AI-driven medical devices. These regulations collectively impact the deployment of generative AI in healthcare by ensuring legal accountability, protecting patient data, and promoting ethical standards in AI development and application.

The implementation of regulatory frameworks such as the EU’s AI Act, HIPAA, GDPR, and MDR significantly impacts the deployment of LLMs and generative AI in healthcare by ensuring transparency, data protection, and patient safety. These regulations necessitate detailed documentation of AI models, advanced data encryption, strict access controls, and rigorous clinical testing, thereby increasing development costs and timelines. However, they also promote reliability, legal accountability, and ethical standards in AI development, fostering trust among users and stakeholders and encouraging the responsible and wider adoption of AI technologies in healthcare \cite{hacker2023regulating}.

\section{Conclusion}
In conclusion, the integration of large language models (LLMs) in healthcare showcases immense potential for enhancing clinical language understanding and medical applications. These models offer versatility and sophistication, from named entity recognition to question-answering, bolstering decision support and information retrieval. Comparative analyses of state-of-the-art LLMs and open-source options emphasize their significance in healthcare, promoting innovation and collaboration. Performance metrics drive continuous improvement but call for rigorous evaluation standards, considering potential biases and ethical concerns. However, challenges persist, including the need for robust training data, bias mitigation, and data privacy. LLMs in healthcare necessitate further research and interdisciplinary cooperation. LLMs promise transformative benefits, but their full potential hinges on addressing these challenges and upholding ethical standards. The ongoing journey of LLMs in healthcare demands collective efforts to harness their power for improved patient care while ensuring ethical and responsible application.


\bibliographystyle{unsrt}  
\bibliography{references}

\end{document}